\documentclass[smallextended]{svjour3}       
\usepackage{graphicx}
\usepackage{times}
\usepackage{epsfig}
\usepackage{graphicx}
\usepackage{amsmath}
\usepackage{amssymb}
\usepackage{url}
\usepackage{multirow}
\usepackage{algorithmicx}
\usepackage[ruled]{algorithm}
\usepackage{algpseudocode}

\usepackage[pagebackref=true,breaklinks=true,letterpaper=true,colorlinks,bookmarks=false]{hyperref}

\journalname{Soft Computing}
\begin{document}

\title{Recognizing Gender from Human Facial Regions using Genetic Algorithm}


\author{Avirup Bhattacharyya 
	\and
Rajkumar Saini 
\and
Partha Pratim Roy \and
Debi Prosad Dogra \and
Samarjit Kar}


\institute{Avirup Bhattacharyya \\
						\email{avirupiem@gmail.com} \\
						Department of Electronics and Communication Engineering, Institute of Engineering and Management, Kolkata, India.\\
						Rajkumar Saini\\
						\email{rajkr.dcs2014@iitr.ac.in}\\ 
              Department of Computer Science and Engineering, Indian Institute of Technology, Roorkee, India.\\
              Partha Pratim Roy\\
              \email{proy.fcs@iitr.ac.in}\\
              Department of Computer Science and Engineering, Indian Institute of Technology, Roorkee, India.\\
              Debi Prosad Dogra\\
              \email{dpdogra@iitbbs.ac.in}\\
              School of Electrical Sciences, Indian Institute of Technology, Bhubaneswar, India.\\
              Samarjit Kar\\
              \email{samarjit.kar@maths.nitdgp.ac.in}\\
              Department of Mathematics, National Institute of Technology, Durgapur, India.\\
					}

\date{Received: date / Accepted: date}

\maketitle

\begin{abstract}
Recently, recognition of gender from facial images has gained a lot of importance. There exist a handful of research work 
 that focus on feature extraction to obtain gender specific information from facial images. However, 
   analyzing different facial regions and their fusion help in deciding the gender of a person from facial images. In this paper, we propose a new approach to identify gender from frontal facial images that is robust to background, illumination, intensity, and facial expression. In our framework, first the frontal face image is divided into a number of distinct regions based on facial landmark points that are obtained by the Chehra model proposed by Asthana et al. The model provides 49 facial landmark points covering different regions of the face, e.g. forehead, left eye, right eye, lips. Next, a face image is segmented into facial regions using landmark points and features are extracted from each region. The Compass LBP feature, a variant of LBP feature, has been used in our framework to obtain discriminative gender specific information. Following this, a Support Vector Machine based classifier has been used to compute the probability scores from each facial region. Finally, the classification scores obtained from individual regions are combined with a genetic algorithm based learning to improve the overall classification accuracy. The experiments have been performed on popular face image datasets such as Adience, cFERET (color FERET), LFW and two sketch datasets, namely CUFS and CUFSF. Through experiments, we have observed that, the proposed method outperforms existing approaches.

\end{abstract}

\section{Introduction}
\label{intro}
The face of a human being reveals useful information about his/her identity, ethnicity, age, gender, expression, and emotion. Gender identification from facial images plays an important role in various computer vision based applications. It is possible for human beings to correctly determine gender by looking at the facial appearance. The advancements of computer vision technologies have inspired the researchers to design systems capable of performing similar functions \cite{jain2004integrating,rai2010gender,tamura1996male}. Human computer interaction system, surveillance system, content based indexing and searching, biometric, and targeted advertising are some of the areas where it has been widely used. In a typical face recognition system, researchers face a number of challenges due to variations in pose, illumination, occlusion, expression, and lower resolution that essentially hamper a system's performance. Hence, the challenge is to design a system that is fairly invariant to changes in illumination, pose, resolution, and facial expression. These challenges have been effectively dealt with by the authors in \cite{moghaddam2002learning,rai2014gender,zhang2009face}. However, gender recognition in such varying conditions has not been fully addressed yet. This paper deals with gender recognition of frontal face images with variations in illumination, pose, facial expression and image resolution.

Feature dimension and computational complexity are important aspects that a typical gender recognition system must satisfy. Systems with lower feature dimension though quick in time, however, such systems perform poorly. Therefore, a method that offers good classification accuracy with a moderate complexity of computation, can be of good use. The key to any gender recognition problem lies in identifying facial regions that contain reasonably good gender specific information. Many feature extraction methods have been proposed over the years for gender recognition problems. Some of the earliest feature descriptors like simple Local Binary Patterns (LBP) \cite{ojala1996comparative} works with raw pixel intensities and provides less discriminative information essential for gender recognition. However, none of the previous methods adopted for gender recognition focuses on facial region extraction for gender specific information. The proposed framework accomplishes that in this aspect. It is based on extraction of facial region for extracting  gender specific information. This paper demonstrates that existing feature descriptors such as LBP \cite{ojala1996comparative} and its variants can be used to improve the face gender recognition task considering facial regions. Although researchers have focused on gender recognition of infrared and visible light images, gender recognition from sketch images has received less attention.

\begin{figure}[!h!t!b]
	\begin{center}
		\includegraphics[width=1.0\linewidth]{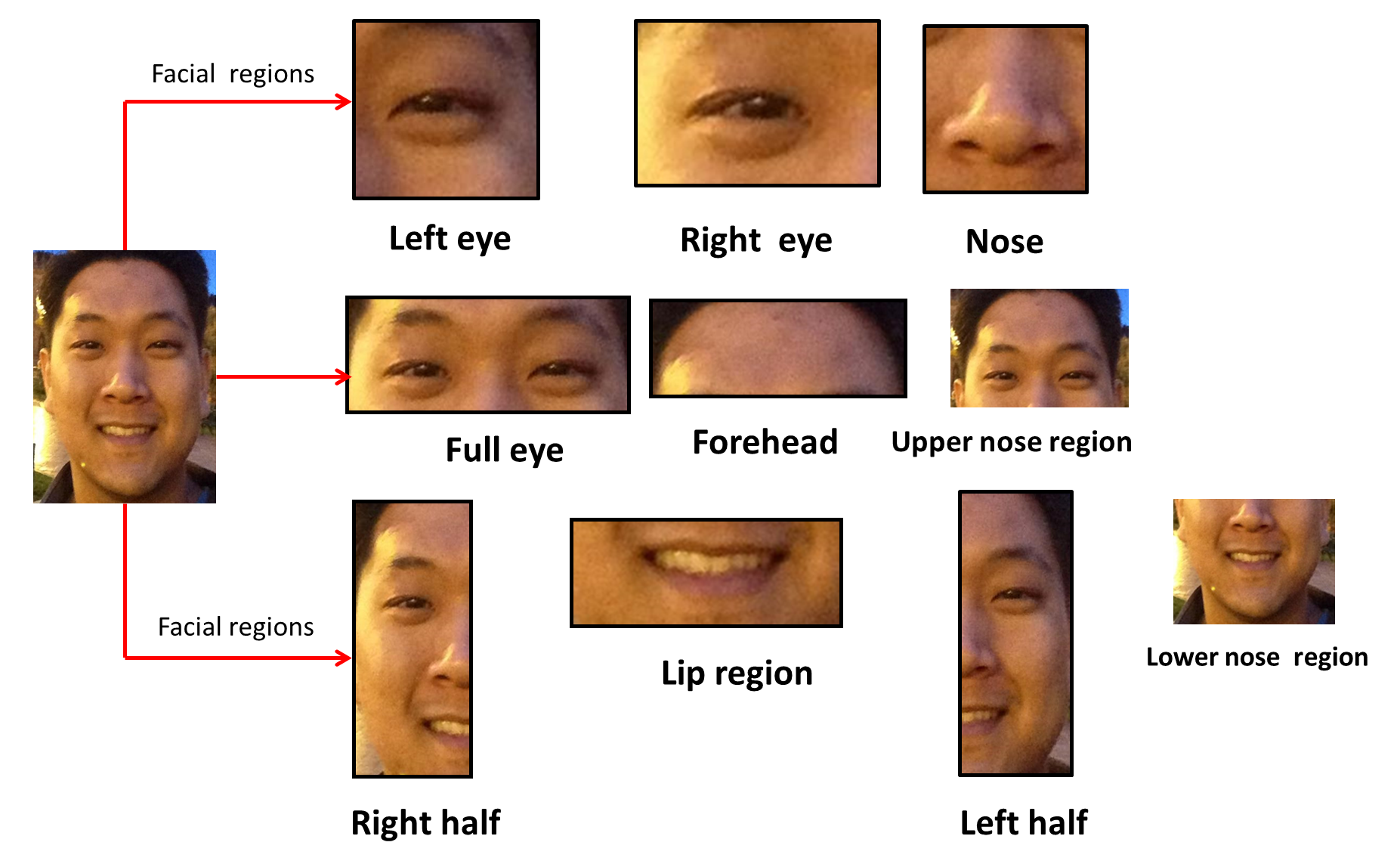}
	\end{center}
	\caption{Example of facial regions that are likely to contribute in predicting the gender of a person.}
	\label{fig:FaceRegions}
\end{figure}

Facial regions can be analyzed separately and their responses can be fused to correctly classify the gender of a person. The facial regions could be more tolerant to variations in expression than the whole face image. This is helpful for gender classification. Fig. \ref{fig:FaceRegions} shows the different facial regions of a  person that have been used in this work for the purpose of gender classification. The proposed method first converts an image into grayscale format and then divides it after detection using Viola Jones algorithm \cite{viola2001rapid} into different sub-images such as eyes, nose, lips, forehead, etc. based on facial landmark points obtained using the Chehra model proposed in \cite{asthana2014incremental}. This is followed by feature extraction from each sub image using Compass Local Binary patterns as used in \cite{patel2016compass}. The most relevant features have been selected using Infinite Feature Selection \cite{roffo2015infinite}. The classification scores of each individual block has been obtained with the help of Support Vector Machine based classifier. Next, classification scores are combined with the help of genetic algorithm. 

The contributions of the paper are as follows; (i) We propose a method of dividing a facial image into different regions to obtain gender specific information from each of them using coLBP. (ii) Region based gender classification has been performed with Support Vector Machine based classifier and corresponding probability scores and classification accuracy from each region has been obtained. (iii) A genetic algorithm based  method for combining the classification scores of individual regions has been proposed. (iv) The method has been compared against a number of state-of-the art methods and it has yielded better results compared to these existing methods. Classification performance has also been compared on facial sketch images and it also provides better results as compared to existing approaches.

The rest of the paper is organized as follows. In Section \ref{sec:relwrk}, we discuss existing work developed for gender recognition using facial images. In Section \ref{sec:proposed}, the proposed framework for region based feature extraction from facial images  and gender classification have been described. Section \ref{sec:resultss} demonstrates the results obtained on various facial image and sketch datasets. In Section \ref{sec:conc}, concluding part of the paper has been described.

\section{Related Work}
\label{sec:relwrk}
The existing works in literature have explored a variety of techniques for feature extraction keeping in mind the challenges mentioned earlier. In \cite{golomb1990sexnet}, authors have used a method which works with raw pixel intensities for feature extraction and subsequently used a neural network for gender identification from images with low resolution. Tamura et al. \cite{tamura1996male} have also used neural networks  for gender identification of low resolution images. Some of the earlier techniques \cite{moghaddam2002learning} recognized gender from images of considerably low resolution using a Support Vector Machine based classifier after using raw pixel intensity based feature extraction techniques. A different kind of approach has been proposed by Baluja and Rowley \cite{baluja2007boosting} where Adaboost classifier has been used for gender classification of low resolution images. Prior to the classification step, feature extraction has been performed using various types of operators for pixel comparison.

The LBP operator has been originally used for texture classification by Ojala et al. in \cite{ojala1996comparative}. The work in \cite{ahonen2006face} proved that Local Binary Patterns can be used for face description. Since LBP works with raw pixel intensities of images, it is less effective in illumination variances. Some of these difficulties have been overcome in \cite{zhang2009face}, where gradient values are computed to provide better edge information. Recently, Precise Patch Histograms (PPH) obtained from the Active Appearance Model (AAM) have been used for gender recognition \cite{shih2013robust}. The authors in \cite{rai2010gender} have used a method involving Radon and Wavelet transforms to extract global face features invariant to illumination changes. They have evaluated the performance of their method on the SUMS dataset.  Following this, classification has been done using Support Vector Machine based classifier (SVM). They have achieved considerably good results. The method in \cite{xie2010fusing} tries to encode gabor phase information by using Local Gabor XOR patterns (LGXP). It also handles illumination changes. The work done in \cite{patel2016compass} is quite tolerant to variations in illumination. The researchers in \cite{Zhang2005} have also handled lighting variations with the help of a pattern named local gabor binary pattern histogram sequence.

However, determination of gender from partially occluded face images has not been studied widely.
Very recently, the authors in \cite{rai2014gender} have dealt with the design of a system robust to occlusion by subdividing a face image into a number of sub images and used gabor filter based 2D Principle Component Analysis for feature extraction. Li et al. \cite{li2012gender} have identified  six regions of the face such as nose, mouth, hair, eyes, forehead, and clothing  that can be used for analysis. This has proved to be useful in classifying gender.

The work done by Tan and Triggs in \cite{tan2010enhanced} proposes a different kind of feature descriptor other than LBP. It is known as Ternary Local Pattern (TLP). It is more robust to noise than LBP due to ternary encoding. Another approach known as Extended Local Binary Pattern  \cite{Huang2007} is more effective than LBP. Some advanced approaches based on LBP  such as Local Directional Pattern \cite{jabid2010gender}, Enhanced Local Directional Pattern \cite{zhong2013face}, eight Local Directional Pattern \cite{faraji2015face} have also been tried by various research groups. These methods perform convolution of the image with different types of filters, better known as Kirsch compass masks. This calculates the edge response images. This provides better structural information and enhances the discriminative power.

A method named Pixel pattern based texture feature (PPBTF) has been adopted by Lu et al. \cite{lu2008automatic} for the task of gender classification. For classification, Adaboost and SVM classifier has been used to achieve appreciably good results. Authors of \cite{lyons2000classifying} have extracted features from images by using gabor filter. They have used Principle Component Analysis (PCA)  for reducing feature dimension. This was followed by Linear Discriminant Analysis (LDA) for classification purpose.  
Andreu and Mollineda in\cite{andreu2008complementarity} have used gray level linear vectors to represent sub images of face and subsequently applied PCA to reduce feature dimension  in order to boost the information. Jain and Huang \cite{jain2004integrating} have used Independent Component Analysis (ICA) along with SVM  for gender identification purpose.

Ozbudak et al. \cite{ozbudak2010fast} have 
used Discrete Wavelet transform  on facial images, followed by PCA for reducing image dimension and finally Fisher Linear Discriminant has been applied to get an accuracy around 93.4\%. In \cite{berbar2014three}, the authors have tried out different feature extraction techniques such as the Discrete Cosine Transform (DCT), Wavelet Transform, Gray-Level Co-occurrence matrix (GLCM). Following this, classification has been done using Support Vector Machine based classifier (SVM). They have achieved considerably good results. Some approaches have also dealt with pose variations. The experiments in \cite{makinen2008evaluation} showed the importance of aligned faces on the classification performance in gender recognition problems. The accuracy obtained is about 87.1\% and the databases used for evaluation are IMM and FERET.The work in \cite{zheng2011support} have used a combination of different methods for feature extraction followed by SVM for classification. They have checked the performance of their method on the CAS-PEAL dataset for multi view images and FERET dataset for frontal face images. For classification purposes, they have combined various linear classification algorithms.In another method \cite{bekios2011revisiting}, it has been shown that evaluating performance on a single database may not yield correct results. As a result, they have included training samples from more than one database.This method also confirmed the effect of age and pose on gender classification results and proved that the robustness of the classifier can be enhanced by exploiting these dependencies.

The methods adopted in \cite{perez2012gender} and \cite{tapia2013gender} have used an approach based on entropy (mutual information). The authors have eliminated redundant information by performing feature selection and next, feature concatenation has been performed to achieve good results. There have been many approaches adapted earlier which aimed at gender classification in both constrained and unconstrained environments. The most recent approach \cite{patel2016compass} involves dividing the whole facial image into $N \times N$ grids and then computing the compass LBP histogram for each grid followed by concatenation of those histograms. One important drawback of this method is that, the arbitrary division of a facial image into $N\times N$ blocks does not provide good local information. In this paper, we propose an improved system  based on facial region extraction that outperforms existing gender recognition systems that use facial images. The results have been evaluated on various databases like LFW, ADIENCE, and cFERET and proposed method works better as compared to existing approaches of gender recognition. The method works reasonably well on frontal face images with variations in pose, illumination, intensity, and expression.

\section{Proposed Framework}
\label{sec:proposed}
In the proposed method, a facial image is first divided into a number of regions using the Chehra model used in \cite{asthana2014incremental}. We have obtained 49 facial landmark points covering the entire face image. Prior to obtaining the landmark points, face detection is carried out using Viola Jones face detection algorithm \cite{viola2001rapid}. These landmark points are then used to obtain facial regions.The method of landmark detection and region extraction have been discussed in detail in Section \ref{sec:subsection1}.  Following this, we have extracted features from facial regions of the images using Compass LBP \cite{patel2016compass} by dividing them into grids of different dimensions e.g. 2$\times$2, 3$\times$3 and 4$\times$4. Of the large number of features, most relevant features from each region have been considered by using a feature selection technique, known as Infinite Feature Selection\cite{roffo2015infinite}. Feature extraction and selection process are discussed in Sections \ref{sec:dwt} and \ref{sec:ifs}, respectively. 
Extracted features are then fed to an SVM classifier. This has been explained briefly in section \ref{sec:svm}. The classifier returns classification scores of all different regions for a particular facial image. These scores are then combined with a genetic algorithm based approach.This has been explained in section \ref{sec:gaa}. The steps are illustrated with the help of a diagram shown in Fig. \ref{fig:Framework}. 

\begin{figure} [!h!t!b]
	\centering
	\includegraphics[width=0.45\linewidth]{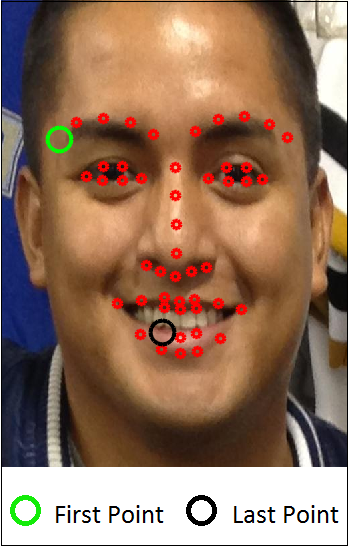}
	\caption{A sample image marked with the landmark points.}
	\label{fig:fig16}
\end{figure}

\begin{figure}[!h!t!b]
	\begin{center}
		\includegraphics[width=1\linewidth]{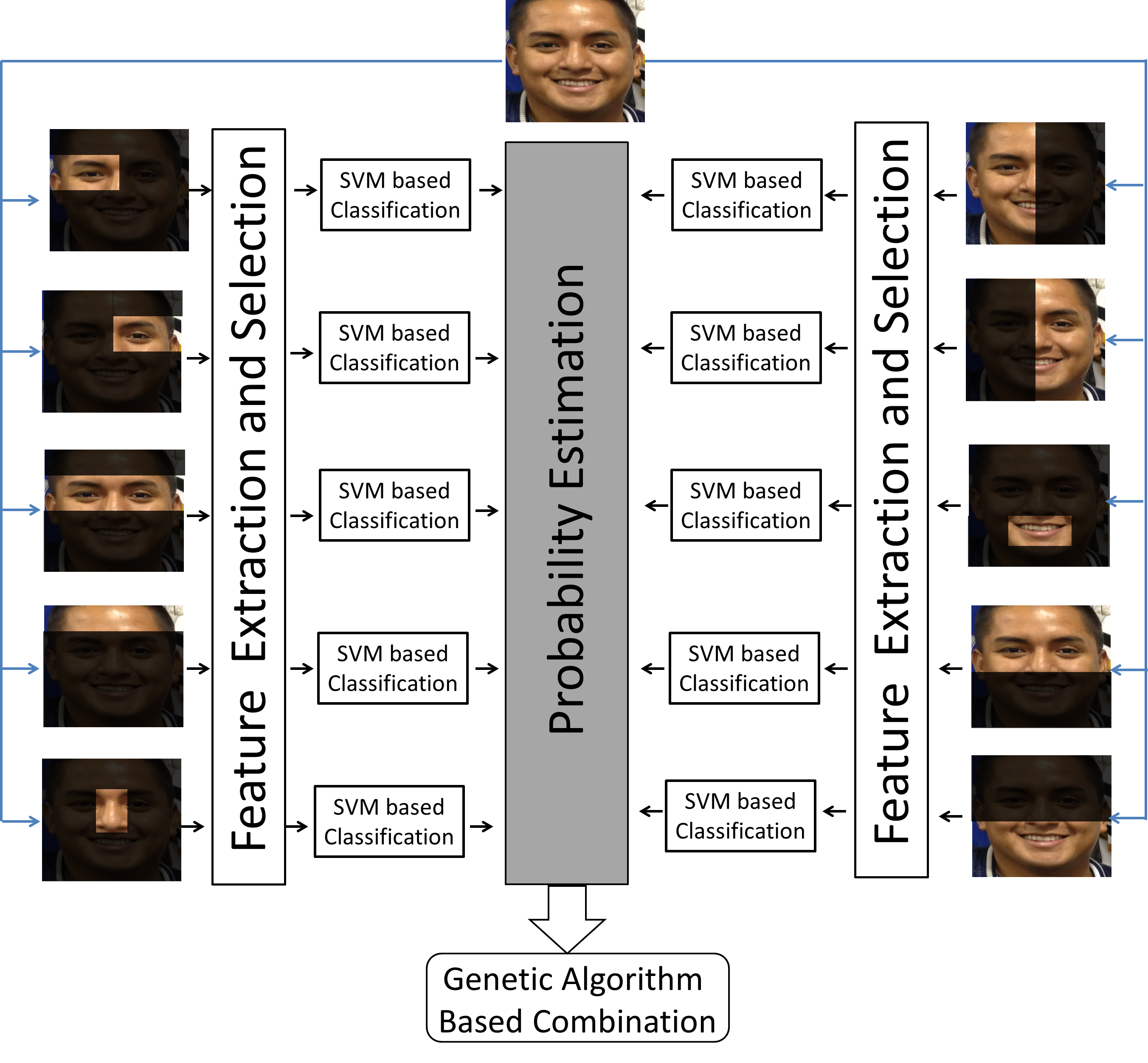}
	\end{center}
	\caption{The block diagram of proposed framework. The features are extracted from the facial regions using Compass LBP followed by Infinite Feature Selection \cite{roffo2015infinite}. The gender specific information obtained from different facial regions have been subjected to SVM based classification. This has followed by genetic algorithm based score combination.}
	\label{fig:Framework}
\end{figure}

\subsection{Facial Landmark point detection and region extraction}
\label{sec:subsection1}
In our work, facial landmark point detection has been done with the help of Chehra model developed in \cite{asthana2014incremental}. The task of constructing and aligning general deformable models capable of obtaining the variations of a non-rigid object such as human face, is an interesting problem in computer vision. On the basis in which the deformable models are constructed and their respective alignment procedure, methods are groups into two types: generative method and discriminative method. The main challenge for the deformable model is the problem encountered in modeling facial texture. The generative model of the sparse facial shape is advantageous in a sense that it can be trained on faces captured in constrained environment to represent shapes of completely different facial images captured in unconstrained environments. As a result, authors in \cite{asthana2014incremental} have focused entirely on incremental updation of the function that maps facial texture to facial shape. The Chehra model computes 49 facial landmark points on the whole facial image. It is actually a discriminative facial deformable model which has been obtained by training a cascade of regressors. The order in which all the 49 facial points are computed is essential for facial region extraction. This has been demonstrated in Fig. \ref{fig:fig16}.

Following landmark point detection, we compute 10 regions on the whole facial image using \cite{asthana2014incremental}. The regions are left eye region, right eye region, complete eye region, lower nose region, lip region, left face region, right face region, forehead region and upper nose region. After obtaining the blocks, feature extraction has been performed with Compass LBP as explained in Section \ref{sec:dwt}. 

\subsection{Feature Extraction using Compass LBP}
\label{sec:dwt}
One of the popular feature descriptors used for image classification problems is LBP, originally used in \cite{ahonen2006face}. It works by comparing the intensity of a center pixel against the intensity of a set of $n$ neighboring 
 pixels lying on the circumference of a circle of radius $r$. If the intensity of the center pixel is more than the neighboring 
  pixels, it is encoded as 0, otherwise 1. Next, the $n$-bit binary encoded number is converted to decimal form. In this work, we have considered a neighborhood 
   of size 8. 
    A uniform LBP is one that contains at most two bit-wise transitions 
     in the binary encoded scheme. The feature dimension of uniform LBP is thus less than the original LBP scheme. The feature dimension of uniform LBP is 59 as it contains only the uniform patterns out of all possible 256 LBP patterns for 8 bit scheme. Mathematically, LBP operator can be expressed as given in (\ref{eq:lbp}).

\begin{equation}
\label{eq:lbp}
LBP_r(P)=\sum_{p=0}^{|P|-1}g(\phi_p-\phi_c)2^p
\end{equation}

where $\phi_c$ and $\phi_p$ represent intensity of center and $p^{th}$ neighboring pixels and $r$ is the neighborhood (chosen to be 1 in this work). The binary function $g(\eta)$ can now be expressed as: 

\begin{equation}
g(\eta)=\begin{cases}1 & \eta \geq0\\0 & \eta < 0\end{cases}
\end{equation}

\begin{figure}[!h!t!b]
	\centering
	\includegraphics[width=0.7\linewidth]{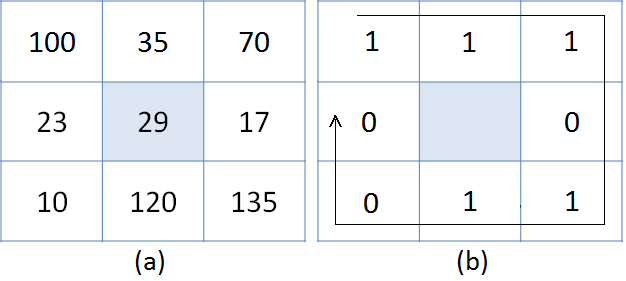}
	\caption{(a) A $3\times3$ window of an image. (b) Equivalent LBP code.}
	\label{fig:exLBP}
\end{figure}

Fig. \ref{fig:exLBP}(b) shows an example of LBP encoding of a $3\times3$ block depicted in Fig. \ref{fig:exLBP}(a). The LBP constructs the code in clockwise manner in the neighborhood of center pixel (here, having intensity 29) starting from top left. Hence, an 8-bit binary number 11101100 is generated for this example. 

\begin{figure}[!h!t!b]
	\begin{center}
		\includegraphics[width=0.7\linewidth]{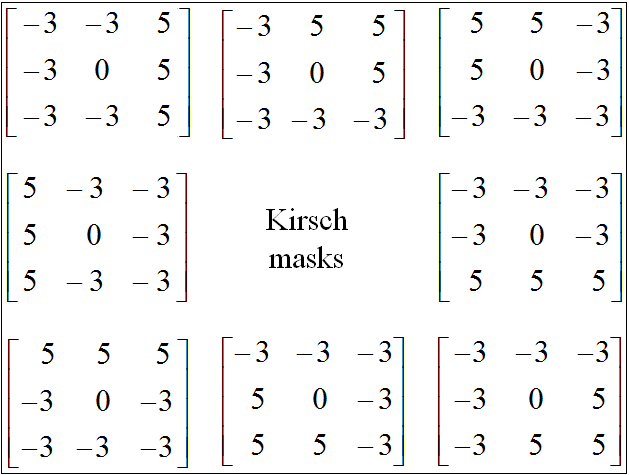}
	\end{center}
	\caption{The set of Kirsch masks used in our work. }
	\label{fig:Kirsch}
\end{figure}

In proposed work, feature extraction has been done with compass LBP as originally proposed and used in \cite{patel2016compass}. It is better than normal LBP as LBP works only with raw pixel intensities. As a result, it is less immune to noise and provides less discriminative information. For computing compass LBP, we have used a set of Kirsch masks that perform image filtering and compute the edge response of each facial region obtained in Section \ref{sec:subsection1} in eight different directions. This enhances the discriminating power. In this method, each facial region is first convolved with all eight Kirsch compass masks shown in Fig. \ref{fig:Kirsch}. After this, the edge responses in all eight directions along with their signs are considered for encoding in binary format using LBP. This is then converted to decimal form. The LBP histograms obtained from each of the eight convolved images for all facial regions have been concatenated to produce the CoLBP histogram. 
 Arbitrary division of a facial image into different blocks may not maintain the perfect locality of information. Hence, it is better to divide the image into discrete blocks such as eyes, nose, lips, forehead, etc. 

%
%
%
%


\subsection{Infinite Feature Selection}
\label{sec:ifs}
Classification based problems in machine learning generally consist of a large feature space involving large number of features that may contain redundant or unwanted data. This results in poor classification performance and also makes the system computationally expensive. Hence, it becomes necessary to select a feature subset that is highly relevant to the problem to improve the performance of classification. In our framework, we have used Infinite Feature Selection  
 (IFS) \cite{roffo2015infinite} for feature selection. It is an unsupervised feature selection technique that has outperformed many of the existing  techniques. It is a graph based approach that uses the convergence property 
  to determine the relative importance of one feature with respect to others. In this method, the features are mapped onto an affinity graph, where features are represented by nodes and weighted edges represent the relationships between them. Each and every  path of a specific length over the graph represents a possible selection of features. Varying these paths and letting them move towards infinity allows us to determine the importance of each possible subset of features. The method provides a final score to each and every feature of the initial set of features. This score gives an indication of the efficiency of the feature in the classification task. The higher the weight of a feature, more it is important in the classification task under consideration. Thus, the most significant feature has the highest weight and the least significant feature has the least weight. Thus,ranking in descendant order the outcome of the Infinite Feature Selection allows us to perform a subset feature selection through a model selection stage to determine the number of features to be selected. 
 

\subsection{Classification using Support Vector Machine}
\label{sec:svm}
After selecting the most relevant features from each facial region, we have computed the classification accuracy of each region. For this purpose, we have used a support vector machine (SVM) \cite{chapelle2002choosing} based classifier. It is a supervised classifier that tries to obtain the best hyper plane (from a set of possible hyper planes) that separates the data points of two classes with the maximum margin. Each training sample $\{x_i,y_i\}$ for $i=1,..., m$ and $y_i \in (-1,1)$ must satisfy (\ref{eq:svm1}) and (\ref{eq:svm2}),
\begin{equation}
\label{eq:svm1}
wx_i+b\geq +1\ \ for\ \ y_i=+1  
\end{equation}
\begin{equation}
\label{eq:svm2}
wx_i+b\leq -1\ \ for\ \ y_i=-1  
\end{equation}
where $w$ and $b$ represent hyperplane and offset, respectively. SVM finds a decision boundary by maximizing the distance between the two parallel hyperplanes. The process of finding this decision boundary is done by minimizing $||w||^2$ which is solved using the Lagrange optimization that returns the hyper planes distinguishing among data classes. We have used linear SVM in this work that returns the probability scores for each sample to be in each class. 
%
The probability scores of individual regions are then combined with genetic algorithm based learning to improve the classification performance as shown in the next section. 
\subsection{Combination using Genetic Algorithm}
\label{sec:gaa}
Genetic algorithms \cite{GAGoldberg} are a class of evolutionary algorithms that are widely used in solving both constrained and unconstrained optimization problems in machine learning. It is a soft computing technique that is based on the process of natural selection that emulates biological evolution. It is actually a simulation of Darwin’s theory of survival of the fittest among individuals over consecutive generations for solving a problem. Although genetic algorithms are randomized, they are not all random, instead they direct the search into the region of better performance in the search space. Each generation consists of a population of individuals. Each individual of the population represents a point in the search space and a possible solution to the problem. Every individual is coded as a finite length vector of components or variables, usually in terms of some alphabet. To maintain the analogy to Darwin’s theory, a solution is analogous to a chromosome and the variables that constitute them are analogous to genes. The algorithm consists of a population of chromosomes with associated fitness values. Based on the fitness values of chromosomes, the algorithm selects individuals to produce offspring’s for the next generation. The fitness value is determined by an objective function. This represents the process of natural selection. The next step after the process of selection is the process of crossover. A crossover site is chosen randomly along the two bit strings. The values of the two individuals are exchanged upto this point. The two new offsprings created from this generation are then put into the next generation of population. Next, mutation can be done by randomly changing few bits of off-springs to tackle the problem of local minima.

\begin{figure}[!h!t!b]
	\centering
	\includegraphics[width=1\linewidth]{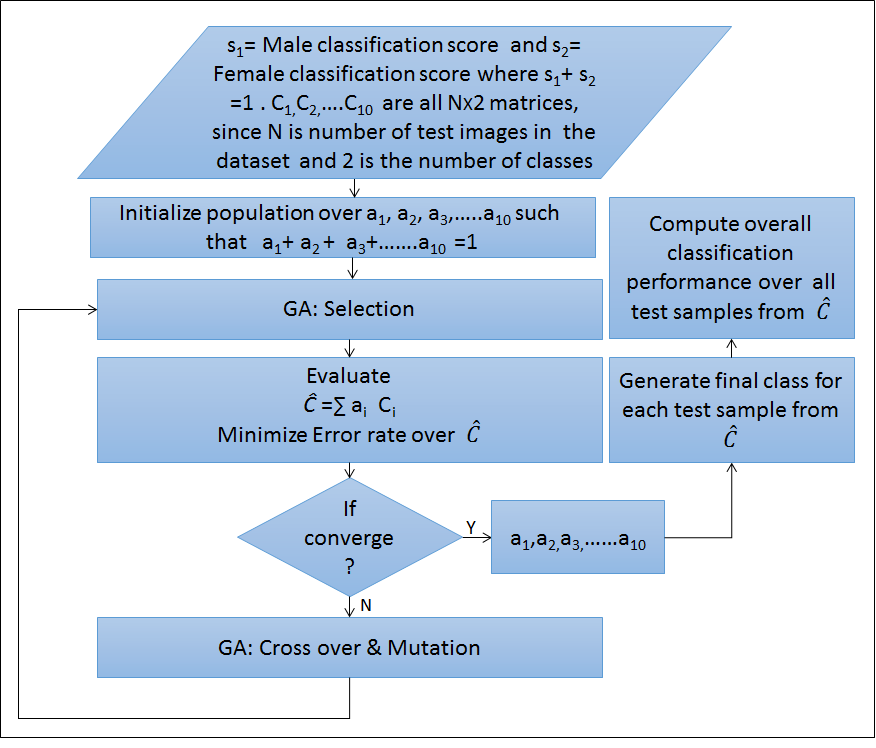}
	\caption{The flowchart of the combination of facial regions using Genetic algorithm.}
	\label{fig:fig45}
\end{figure}

In this paper, we have presented a genetic algorithm based method for combining the classification scores of different facial regions obtained earlier with SVM. The population has been randomly initialized over $a_1$,$a_2$,....$a_{R}$ where  $a_1$,$a_2$,....$a_{R}$ are the weights assigned to individual facial regions. The classification error has been considered as the objective function to minimize. The final class has been generated for each test sample from the linear weighted combination of classification scores. The flowchart of the GA based combination has been shown in Fig. \ref{fig:fig45}. The equation for the weighted linear combination of classification results of different facial regions using their probability scores has been performed using (\ref{eq:GAcombination}). This is the equation that we attempt to optimize in this problem and may be stated as:

\begin{equation}
\hat{C}=\sum_{i=1}^{R}a_iC_i
\label{eq:GAcombination}
\end{equation}

where $a_i$ represents the weight assigned to the $i^{th}$ (i=1,2..$R$) classification block $C_i$. $C_i$ consists of probability scores for $i^{th}$ facial region. Over the iteration GA tries to find the optimal weights to improve the classification accuracy.

The framework has been provided in Algorithm \ref{algo:algorithm2}. The input to the algorithm is the  facial image from the dataset. Step 3 in the algorithm explains the face detection process using Viola Jones algorithm. Steps 4 and 5 explain the process of landmark point detection and region extraction. The next two steps explain in detail the process of feature extraction and feature selection. 
Steps 9 to 14 explain the classification of each facial region and the genetic algorithm based combination.

\alglanguage{pseudocode}
\begin{algorithm}[!h!t!b]
	\small
	\caption{Algorithm for the proposed framework}
	\label{algo:algorithm2}
	\begin{algorithmic}[1]
		\Procedure {$\mathbf{Input:}$}  $Facial$ {$grayscale-image$} {$from$} {$the$} {$ dataset$}
		\For {$i = 1 \to \|{n}\|$}		
		\Comment{n is the total no of images which varies according to the dataset}
		\State Perform face detection using Viola Jones algorithm  \cite{viola2001rapid} to calculate bounding box coordinates $b_1$,$b_2$,$b_3$,$b_4$ for each image.\;
		\State Obtain facial landmark points  $ P_1$, $P_2$, $P_3$, ........$P_{49}$ using {\textbf{Chehra model}} in  \cite{asthana2014incremental}.\;
		\Comment{49 points computed by the Chehra model}.
		\State Extract facial regions $r_1$, $r_2$, $r_3$, $r_4$,.......$r_{10}$  from the landmark points.\;
		\State Compute the features $f_1$, $f_2$, ....... $f_j$ (where j varies depending on the regions formed using mask size such as 2$\times$2) from each region using { \textbf{Compass LBP}} \cite{patel2016compass}.\;
		\State Reduce feature dimension using {\textbf {Infinite Feature Selection}} \cite{roffo2015infinite} to select most relevant features $f_1$,$f_2$,$f_3$,....$f_l$ $(l<j)$ from each $r_i$. 
		\EndFor
		\For {$v = 1 \to \|{10}\|$}
		\For {$i = 1 \to \|{n}\|$}
		\State for $i^{th}$ sample image calculate the classification score $C_v$ of face region $r_v$ using SVM.\;
		\EndFor
		\EndFor
		\State Combine the classification scores {$C_1$, $C_2$,........$C_{10}$} using {\textbf{Genetic Algorithm}} \cite{GAGoldberg}\;
		\State {$\mathbf{Output:}$} {$Gender$} {$identification$} {$of$} {$the$} {$input$}	{$image$}	
		\EndProcedure 
	\end{algorithmic}		
\end{algorithm}

\section{Experiments and Results}
\label{sec:resultss}
In this section, we present the results of the gender classification using public datasets to test the proposed framework. Linear SVM has been used to perform the classification task.
\subsection{Dataset Description}
We have evaluated the performance of the proposed method on three facial image datasets and two sketch datasets. The facial image datasets are Adience, LFW and color FERET. 

\subsubsection{Adience dataset}

\begin{figure}[!h!t!b]
	\begin{center}
		\includegraphics[width=0.6\linewidth]{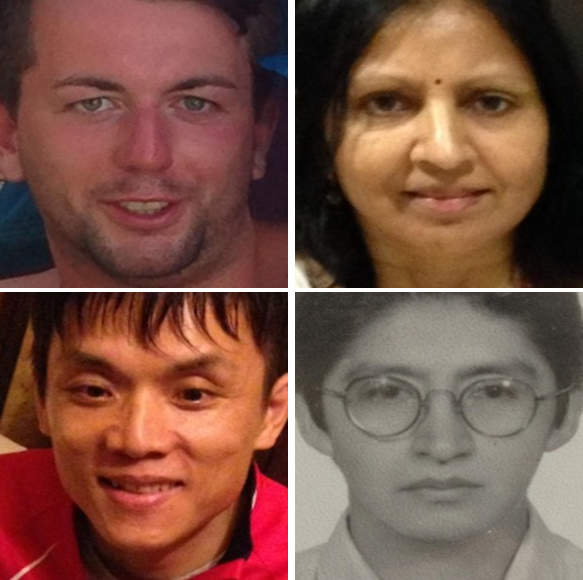}	
	\end{center}
	\caption{Sample images of Adience dataset.}	
	\label{fig:Adience}
\end{figure}

The Adience dataset \cite{AdienceDataseteidinger2014age} contains both frontal and non-frontal facial images of people belonging to different countries, races, and age groups. Their label information is given along with the dataset. In our work, we have considered only frontal face images and ignored the non-frontal images. Some facial images of children where the gender is not clear by visual observation, have not been included in the study. The images contain varying illumination, background, occlusion and expression. A set of 1757 images used by the authors in \cite{patel2016compass} has been used in this work following above protocol. This includes 840 males and 917 females. Fig. \ref{fig:Adience} shows sample images from the Adience dataset.


\subsubsection{LFW dataset}

\begin{figure}[!h!t!b]
	\begin{center}
		\includegraphics[width=0.6\linewidth]{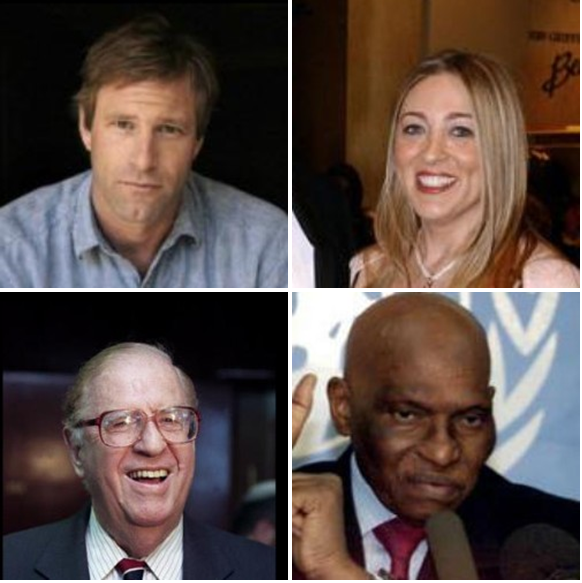}		
	\end{center}
	\caption{Sample images from LFW dataset.}
	\label{fig:lfw}
\end{figure}

The LFW dataset \cite{LFWhuang2007labeled} is also as challenging   
as Adience dataset. It contains approximately 13000 face images of 5749 individuals. The dataset is unlabeled and hence manual labeling of images has been required. There are varying number of images for an individual present in the dataset. However, only one image per person has been considered in our work to avoid biased classification of gender. Thus, a total of 5749 images have been used for experiments in this work. 
  Fig. \ref{fig:lfw} shows a few samples from the LFW dataset.

\subsubsection{Color FERET dataset}

\begin{figure}[!h!t!b]
	\begin{center}
		\includegraphics[width=0.6\linewidth]{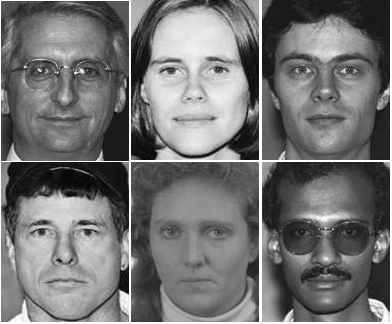}		
	\end{center}
	\caption{Sample images from color FERET dataset.}
	\label{fig:feret}
\end{figure}

The color FERET database \cite{FERETphillips2000feret} is considered as a standard database for evaluating the performance of a given gender recognition system. It consists of a number of images from 994 individuals with varying illumination and expression. However, only one image per individual has been considered in our work. The collection used in our work consists of 987 images out of which 587 are male and 400  are female face images. This collection of 987 images has been selected to maintain parity with the CUFSF dataset that contains the sketches of the facial images in the FERET 
 database. Fig. \ref{fig:feret} shows some samples from the cFERET dataset.

%
\subsubsection{Sketch datasets}
\begin{figure}[!h!t!b]
	\begin{center}
		\includegraphics[width=1\linewidth]{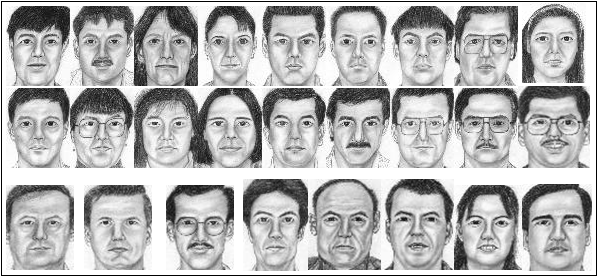}	
	\end{center}
	\caption{Sketch datasets: Samples from CUFSF and CUFS datasets.}
	\label{fig:cufs}
\end{figure}

The sketch datasets considered in our work are CUFS and CUFSF datasets \cite{CUFSandCUFSFwang2009face}. We have considered the same set of 987 images for CUFSF dataset  
 in our work as used in \cite{patel2016compass} to provide a fair idea about comparative performance. The CUFS dataset contains 606 facial sketches of subjects from three different datasets such as the CUHK student database, AR database, and XM2VTS database.  Fig. \ref{fig:cufs} shows a few samples from CUFS and CUFSF datasets.   

\subsection{Performance using individual face regions}
We have carried out experiments on a system having intel core i3 processor and 8 GB of RAM. The method has been evaluated by comparing the classification performance of each facial region  individually and also the performance after combination of all the facial regions. It has been observed that the eye region contains more gender specific information compared to other regions of the face.  The table lists the classification performance of the individual regions for the Adience dataset. Table \ref{tab:adience} shows the accuracy of different regions on Adience dataset.
%

It is evident that the eyes of an individual contains the maximum gender specific information of all the regions while the forehead contains the least.

\begin{table}[!h!t!b]
	\centering
	\caption{Region wise accuracy on Adience dataset using grid size of 4x4 }
	\label{tab:adience}
	\begin{tabular}{|c|c|}
		\hline
		\textbf{Facial Region} & \textbf{Accuracy (\%)} \\ \hline
		Left Eye           & 84.06             \\ \hline
		Right Eye              & 82.93             \\ \hline
		Complete Eye           & 83.27             \\ \hline
		Forehead               & 73.95             \\ \hline
		Lip Region             & 78.25             \\ \hline
		Nose Region            & 77.19             \\ \hline
		Lower Nose             & 79.56             \\ \hline
		Left Half of the Face  & 82.71            \\ \hline
		Upper Nose             & 75.81             \\ \hline
		Right Half of the Face & 80.32             \\ \hline
	\end{tabular}
\end{table}

\begin{figure} [!h!t!b]
	\centering
	\includegraphics[width=1.2\linewidth]{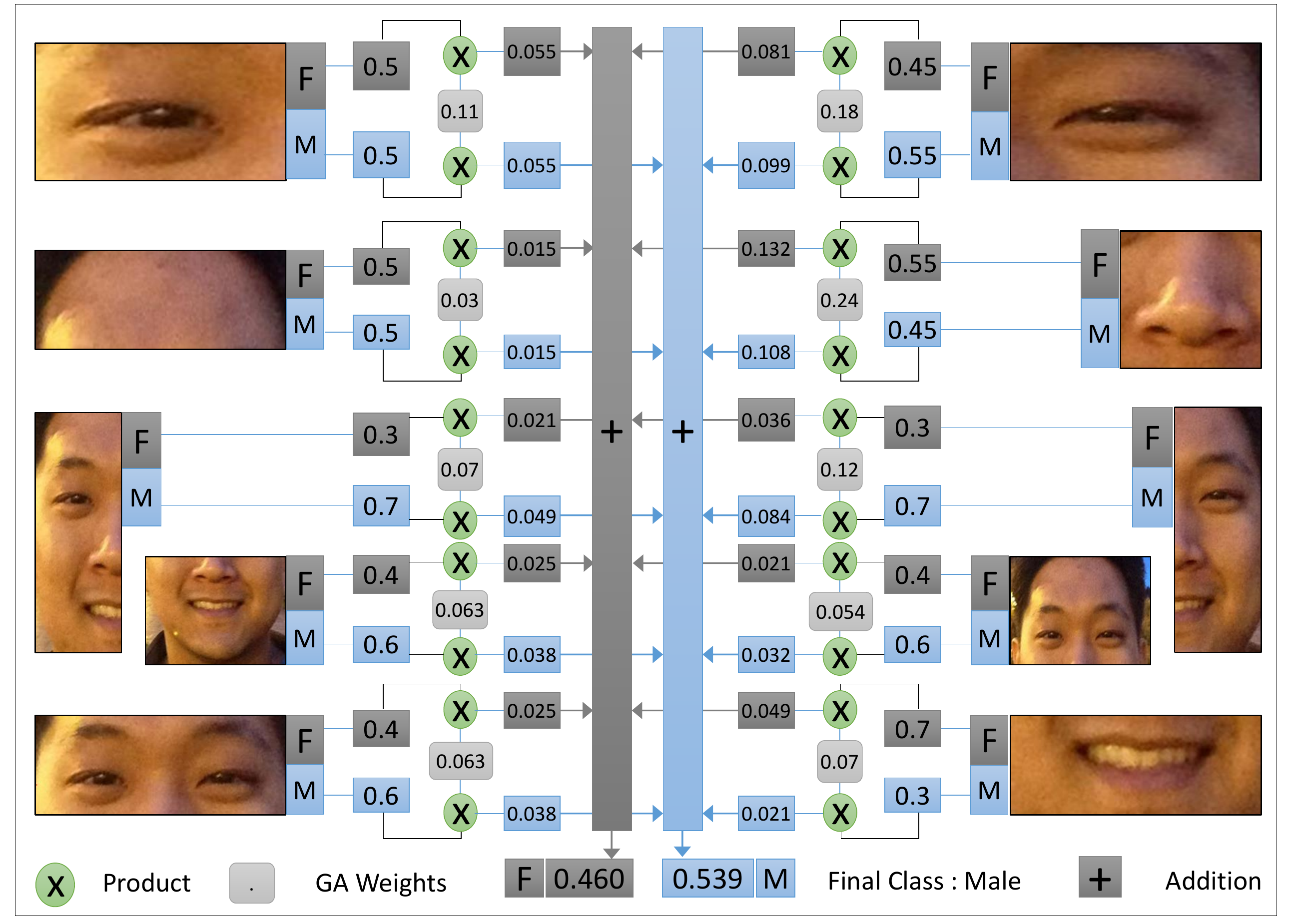}
	\caption{Example of applying GA optimized weights for final gender recognition.}
	\label{fig:41}
\end{figure}

GA has been used for optimization by combining the scores from different face regions. Fig. \ref{fig:41} shows how the final gender has been decided using GA. Weights learned from GA are multiplied with the probability scores and then addition of the weighted scores is performed which results in two scores, one for each gender. The test sample is assigned the gender corresponding to class with the highest score. 

\subsection{Parameter evaluation }
\begin{figure} [!h!t!b]
	\centering
	\includegraphics[width=0.8\linewidth]{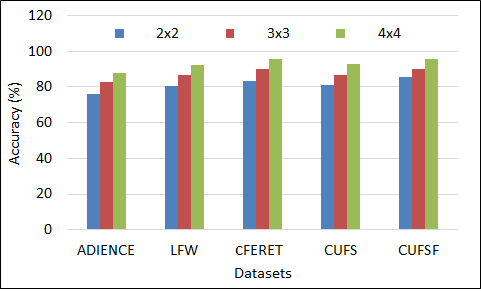}
	\caption{Variation of classification accuracy with grid size for different datasets.}
	\label{fig:fig31}
\end{figure}

In our work, we have considered dividing the facial regions into grids of different sizes such as 2$\times$2, 3$\times$3 and 4$\times$4. The accuracy of each region has  progressively improved with the increase in no of grids for all regions. This can be attributed to the fact that, more relevant features can be extracted with the increase in the number of grids. This helps to obtain better local information.This is particularly true for some of the larger-sized regions. Thus, the best classification accuracy has been obtained with grid size of 4$\times$4. Also, the overall classification accuracy has improved for all datasets with the increase in number of grids (shown in Fig. \ref{fig:fig31}).

\subsection{Evaluation protocol}
The experiments have been carried out using 5-fold cross validation scheme. The dataset is partitioned into five parts. Four of these five parts have been used for training and the other part has been used for testing. This has been carried out five times and each time, a different subset has been used for testing and the rest for training the classifier. 
  This has been followed for all datasets. Male and female classification rates have been determined separately for each dataset. 


We have also shown the confusion matrix  for each dataset as shown in Fig. \ref{fig:fig27}. Fusion of region classification scores has been performed using GA using the parameters as given in Table \ref{tab:GAparams}.

\begin{table}[!h!t!b]
	\centering
	\caption{GA Configuration}
	\label{tab:GAparams}
	\begin{tabular}{|c|c|c|c|}
		\hline
		\textbf{\begin{tabular}[c]{@{}c@{}}Selection\\ Function\end{tabular}} & \textbf{\begin{tabular}[c]{@{}c@{}}Crossover\\ Probability\end{tabular}} & \textbf{\begin{tabular}[c]{@{}c@{}}Mutation\\ Probability\end{tabular}} & \textbf{\begin{tabular}[c]{@{}c@{}}Chromosome\\ Encoding\end{tabular}} \\ \hline
		\begin{tabular}[c]{@{}c@{}}Roulette\\ Wheel\end{tabular}              & 0.80                                                                     & 0.01                                                                    & \begin{tabular}[c]{@{}c@{}}2-bit Double \\ Vector\end{tabular}         \\ \hline
	\end{tabular}
\end{table}


\subsection{Comparative study}
\label{sec:Comparison}
We have compared the performance of our method with different approaches such as neural network based approach, feature concatenation based approach and a few other features exist in literature such as such as ELBP, LGBPHS and CoLBP.

\subsubsection{Comparative study with Neural Network based approach }
\begin{figure}[!h!t!b]
	\centering
	\includegraphics[width=0.8\linewidth]{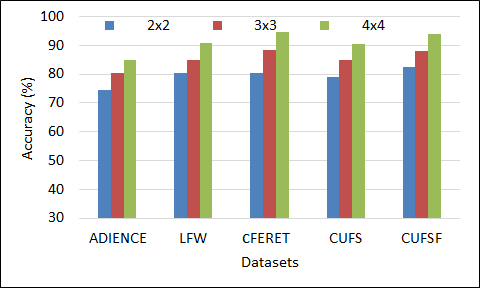}
	\caption{Comparative results of gender recognition when probability scores from SVM are classified using Neural Network.}
	\label{fig:fig32}
\end{figure}

In this section, we present a comparative study with a neural network based classification approach. The neural network has been trained with the probability scores from each of the 10 facial regions obtained using SVM based classifier as done in proposed approach. Thus the neural network has 20 input nodes. The number of nodes in the intermediate `hidden layer' has been taken as 10 from experiments. The number of output nodes is 2 since there are two output gender (male and female). The activation function used is sigmoidal. The  other parameters like learning rate and batch-size have been optimized to obtain maximum classification performance. The study has been carried out for the same combinations of grid size (2$\times$2, 3$\times$3, 4$\times$4) for different datasets. In all cases, our method outperformed the neural network based approach. The results of our observation are presented in Fig. \ref{fig:fig32}. 

\subsubsection{Comparative study with Feature concatenation based approach}

\begin{figure}[!h!t!b]
	\centering
	\includegraphics[width=0.8\linewidth]{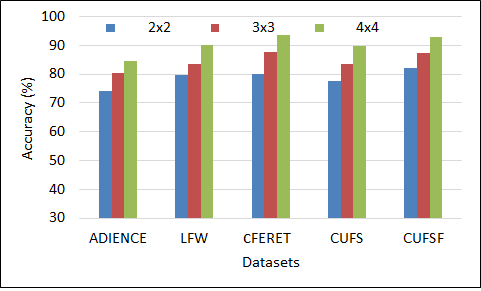}
	\caption{Comparative results of gender recognition with feature concatenation based approach.}
	\label{fig:fig20}
\end{figure}

We present the comparative performance using feature concatenation technique. In this technique, we have concatenated the features from all 10 facial regions (in this way, features from all facial regions share equal weightage) followed by feature selection with Infinite Feature Selection technique to build final feature descriptor. Finally, classification was performed using SVM. Similar to Neural Network based approach, the results obtained with the proposed method are better than feature concatenation based approach on all datasets. The results are presented in the bar graph shown in Fig. \ref{fig:fig20} for grids of size 2$\times$2, 3$\times$3 and 4$\times$4. 


\subsubsection{Comparative study with existing feature extraction methods}

We show the performance of the proposed method using combination of classification results from different facial regions using Genetic Algorithm. The results have been obtained by dividing facial regions into 4x4 grids. Table \ref{tab:CUFSandCUFSF} shows the performance of the proposed system on sketch datasets and compares it  with existing approaches of gender recognition such as ELBP, LGBPHS and CoLBP. The classification accuracy for males and females has been determined separately for all datasets. The overall accuracy has also been computed. In Table  \ref{tab:CUFSandCUFSF}, second column shows the accuracy for males whereas the third column shows accuracy for females. Fourth column represents the overall value of accuracy for CUFS dataset followed by results for CUFSF dataset in same manner. The name of the feature is given in the first column of the table. The accuracy is as high as 92.90\% and 95.85\% using CUFS and CUFSF datasets, respectively. 


\begin{table}[!h!t!b]
	\centering
	\caption{Performance on CUFS and CUFSF sketch Datasets}
	\label{tab:CUFSandCUFSF}
	\begin{tabular}{|c|c|c|c|c|c|c|}
		\hline
		& \multicolumn{3}{c|}{\textbf{CUFS (\%)}}            & \multicolumn{3}{c|}{\textbf{CUFSF (\%)}}           \\ \hline
		\textbf{Method}                                                    & \textbf{Male} & \textbf{Female} & \textbf{Overall} & \textbf{Male} & \textbf{Female} & \textbf{Overall} \\ \hline
		ELBP \cite{Huang2007}                                                              & 92.67         & 84.8            & 89.6             & 95.06         & 87.5            & 92               \\ \hline
		LGBPHS \cite{Zhang2005}                                                             & 94.03         & 83.08           & 89.76            & 96.6          & 89              & 93.52            \\ \hline
		CoLBP \cite{patel2016compass}                                                             & 95.39         & 84.81           & 91.25            & 96.25         & 92              & 94.53            \\ \hline
		\textbf{\begin{tabular}[c]{@{}c@{}}Proposed\\ Method\end{tabular}} & 97.02         & 86.50          & \textbf{92.90}   & 97.96          & 92.75           & \textbf{95.85}     \\ \hline
	\end{tabular}
\end{table}


Similarly, Table \ref{tab:ALF} shows the performance of proposed system on ADIENCE, LFW and cFERET datasets and compares it with existing features such as ELBP, LGBPHS and CoLBP. The accuracy is as high as 87.71\%, 92.29\% and 95.75\% using ADIENCE, LFW and color FERET datasets, respectively. It may be observed that the proposed system outperforms above mentioned existing features. 
 
\begin{table}[!h!t!b]
	\centering
	\caption{Performance on ADIENCE, LFW and color FERET datasets }
	\label{tab:ALF}
		\begin{tabular}{|c|c|c|c|c|c|c|c|c|c|}
			\hline
			& \multicolumn{3}{c|}{\textbf{ADIENCE (\%)}}         & \multicolumn{3}{c|}{\textbf{LFW (\%)}}             & \multicolumn{3}{c|}{\textbf{color FERET (\%)}}           \\ \hline
			\textbf{Method}                                                    & \textbf{Male} & \textbf{Female} & \textbf{Overall} & \textbf{Male} & \textbf{Female} & \textbf{Overall} & \textbf{Male} & \textbf{Female} & \textbf{Overall} \\ \hline
			ELBP                                                               & 76.61         & 78.63           & 77.69            & 93.52         & 70.07           & 87.48            & 96.6          & 87.75           & 93.01            \\ \hline
			LGBPHS                                                             & 81.67         & 80.04           & 80.82            & 97.07         & 67.3            & 89.4             & 97.11         & 85.75           & 92.5             \\ \hline
			CoLBP                                                              & 84.88         & 82.99           & 83.89            & 95.37         & 73.32           & 89.7             & 97.1          & 89.25           & 93.92            \\ \hline
			\textbf{\begin{tabular}[c]{@{}c@{}}Proposed\\ Method\end{tabular}} & 88.69         & 86.80           & \textbf{87.71}   & 97.70         & 76.78            & \textbf{92.29}   & 98.47
			& 91.75           & \textbf{95.75}  \\ \hline
		\end{tabular}
\end{table}

\begin{figure}[!h!t!b]
	\centering
	\includegraphics[width=0.8\linewidth]{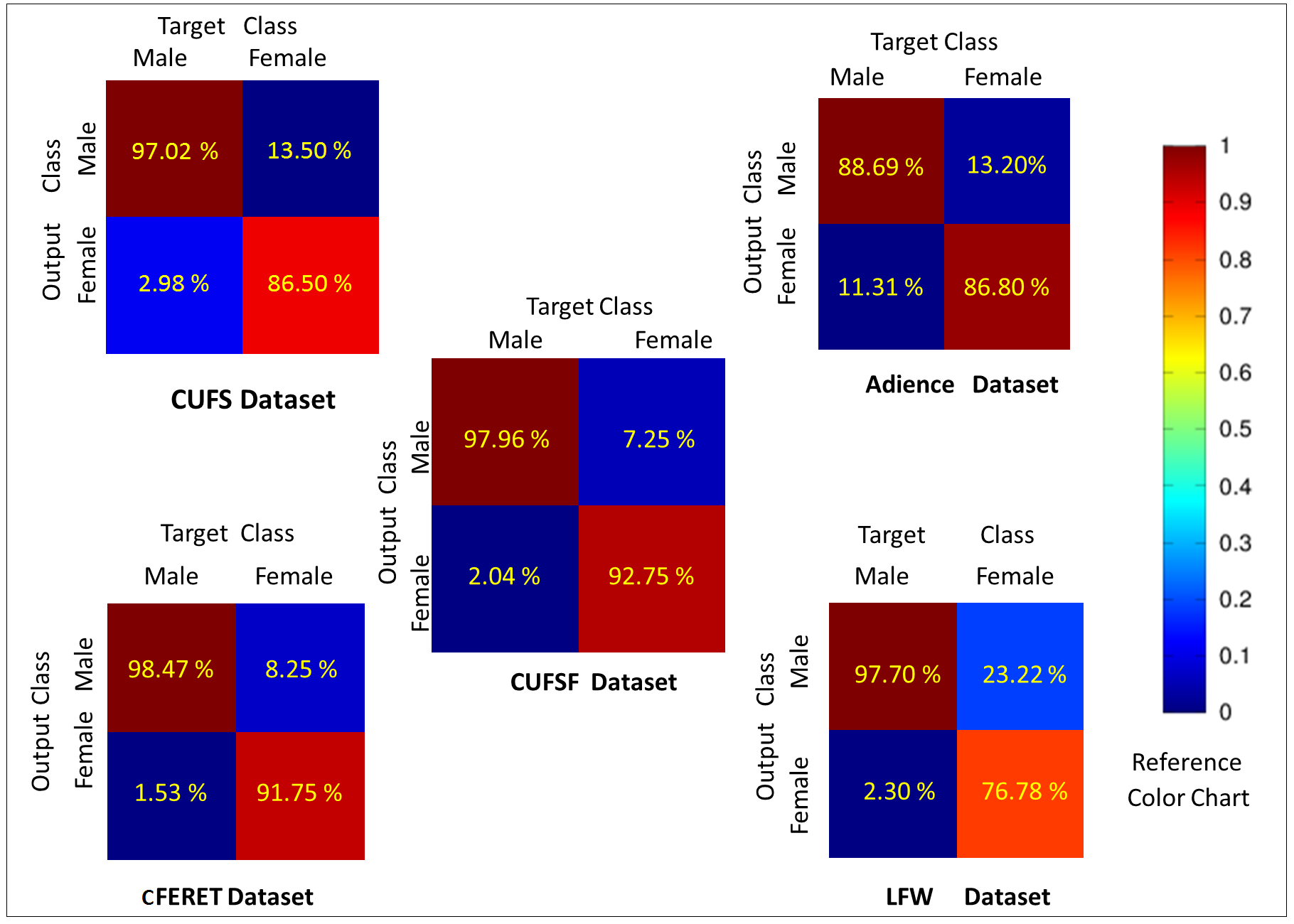}
	\caption{Confusion matrices for all the datasets used in our work with the reference color chart on the right.}
	\label{fig:fig27}
\end{figure}

\begin{figure}[!h!t!b]
	\begin{center}
		\includegraphics[width=0.8\linewidth]{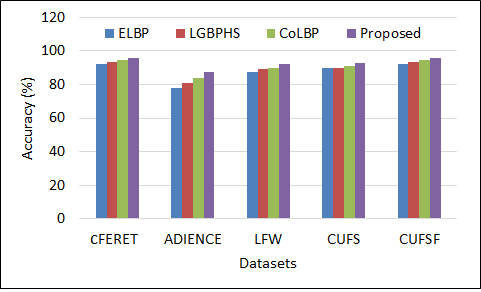}
	\end{center}
	\caption{Performance of proposed framework using Adience, LFW, cFERET, CUFS and CUFSF datasets.}
	\label{fig:results}
\end{figure}

Performance of proposed framework on all the datasets is shown in Fig.\ref{fig:results}. The highest accuracy of 95.85\% has been recorded when evaluated on CUFSF dataset. The classification accuracy has been recorded as high as 95.75\% on color FERET dataset. The accuracy of 87.71\%, 92.29\% and 92.90\% have been recorded using Adience, LFW and CUFS datasets, respectively.  Adience dataset is more challenging due to variations in illumination, resolution and background expression. The confusion matrices of all five datasets are presented in Fig. \ref{fig:fig27}.

The computation overhead has also been analyzed on Adience dataset. Table. \ref{tab:time} shows the computation time of training and testing phases in milliseconds. As is evident from the table, the computation complexity of proposed framework is marginally higher than CoLBP, however, the method can be adopted for efficient gender classification given the higher classification performance.

\begin{table}[!h!t!b]
	\centering
	\caption{Computation overhead on ADIENCE dataset.}
	\label{tab:time}
	\begin{tabular}{|c|c|c|}
		\hline
		& \textbf{\begin{tabular}[c]{@{}c@{}}Training Time\\ (milliseconds)\end{tabular}} & \textbf{\begin{tabular}[c]{@{}c@{}}Testing Time\\ (milliseconds)\end{tabular}} \\ \hline
		ELBP     & 375.5                                                                      & 5                                                                         \\ \hline
		LGBPHS   & 2743.2                                                                     & 50.4                                                                      \\ \hline
		CoLBP    & 610                                                                        & 8.7                                                                       \\ \hline
		Proposed & 723                                                                        & 9.8                                                                      \\ \hline
	\end{tabular}
\end{table}

\section{Conclusion}
\label{sec:conc}
Gender recognition in images is an interesting problem in today’s context. It is widely used in surveillance systems, content based indexing and searching, biometric systems, etc. This paper proposes a new approach to the gender classification task. It divides a facial image into different blocks with the help of facial landmark points obtained using the Chehra model. This is done to obtain accurate local information. As per our study, the method is better than arbitrarily dividing a facial image into different blocks. The feature extraction has been performed with the help of Compass LBP, followed by dimension reduction using Infinite Feature Selection. The classification scores of individual regions have been combined with genetic algorithm to provide improved performance. Compass LBP calculates the edge response of a facial image in 8 different directions with the help of Kirsch masks. Infinite feature selection chooses the most relevant features from each block and thus reduces the computation time and also improves the classification performance. We have compared the proposed method with neural network based approach and feature concatenation based approach. The method has been evaluated on three image datasets such as the Adience, LFW and color FERET. Two sketch datasets, namely CUFSF and CUFS have also been considered for evaluation. The method also works reasonably well on sketch images, which proves its usefulness in crime detection. The computational complexity of the method is comparable with the existing systems. With suitable optimization strategies, the computational complexity can further be reduced and more accurate results can be produced.

{\small
	\bibliographystyle{ieee}
	\bibliography{egbib}
}
\end{document}